\pdfoutput=1

\documentclass[11pt]{article}

\usepackage[preprint]{acl}

\usepackage{multirow}
\usepackage{placeins}
\usepackage{times}
\usepackage{latexsym}
\usepackage{pbox}
\usepackage[T1]{fontenc}

\usepackage[utf8]{inputenc}

\usepackage{microtype}

\usepackage{inconsolata}

\usepackage{colortbl}
\usepackage{graphicx}
\usepackage{xspace}
\usepackage{tabularx}
\usepackage{tabulary}
\usepackage{amsmath}
\usepackage{booktabs}
\usepackage{listings}
\usepackage[most]{tcolorbox}
\usepackage{colortbl}
\usepackage{cuted}
\usepackage{xcolor}
\usepackage{kotex}
\usepackage{pifont}
\usepackage{dirtytalk}
\usepackage{algorithm}
\usepackage{algpseudocode}
\usepackage[utf8]{inputenc}
\usepackage[T1]{fontenc}
\usepackage[perpage]{footmisc}
\usepackage{hyperref}
\usepackage{makecell}

\usepackage{ulem}
\usepackage{makecell}

\usepackage{amsmath, amssymb}
\usepackage[most]{tcolorbox}

\usepackage{geometry}
\usepackage{longtable}
\usepackage{subcaption}

\definecolor{boxbg}{RGB}{245, 245, 245} 
\definecolor{boxborder}{RGB}{100, 100, 100} 

\newtcolorbox{instructionsbox}{
  colback=boxbg,
  colframe=boxborder,
  boxrule=1.5pt,
  arc=4pt,
  left=6pt,
  right=6pt,
  top=6pt,
  bottom=6pt,
  fontupper=\ttfamily, 
}

\newcommand{\model}{\textsc{₩on}\xspace}
\newcommand{\dataset}{\textsc{₩on-Instruct}\xspace}

\title{\model: Establishing Best Practices for Korean Financial NLP}

\author{
Guijin Son{\textsuperscript{1\textsuperscript{*}}} \quad Hyunwoo Ko{\textsuperscript{1}} \quad Hanearl Jung{\textsuperscript{1}} \quad Chami Hwang{\textsuperscript{2}\textsuperscript{†}}
\\ \\ OneLineAI{\textsuperscript{1}} \quad KRX{\textsuperscript{2}} \\ 
\texttt{spthsrbwls123@yonsei.ac.kr} \quad \texttt{hcharm2ing@krx.co.kr} \\ 
}

\begin{document}
\maketitle

\renewcommand{\thefootnote}{\fnsymbol{footnote}}
\footnotetext[2]{Corresponding author.}
\renewcommand{\thefootnote}{\arabic{footnote}}

\begin{abstract}

In this work, we present the first open leaderboard for evaluating Korean large language models focused on finance. Operated for about eight weeks, the leaderboard evaluated 1,119 submissions on a closed benchmark covering five MCQA categories: finance and accounting, stock price prediction, domestic company analysis, financial markets, and financial agent tasks and one open-ended qa task. Building on insights from these evaluations, we release an open instruction dataset of 80k instances and summarize widely used training strategies observed among top-performing models. Finally, we introduce \model, a fully open and transparent LLM built using these best practices. We hope our contributions help advance the development of better and safer financial LLMs for Korean and other languages.

\end{abstract}

\section{Introduction}

Large language models (LLMs) hold significant potential for financial applications~\citep{son2023beyond, chen2024survey, chen-etal-2024-knowledge}. However, performance issues in this domain can lead to monetary losses, making it imperative to develop reliable evaluation systems prior to deployment. Unfortunately, the inherently closed nature of the financial industry limits the sharing of models~\citep{wu2023bloomberggpt} and dataset~\citep{mahfouz-etal-2024-state}, slowing the development of relevant techniques and often resulting in duplicated efforts across companies and teams.

Existing tools to evaluate LLM performance in the Korean financial domain include KRX-Bench~\citep{son2024krx}, which specifically assesses knowledge of Korean listed companies, and KMMLU~\citep{son2024kmmlu}, a broader benchmark spanning 45 categories that incorporates a subset of finance and economics. However, these benchmarks fall short of reflecting the broad potential for LLM applications in the financial sector. To address this gap, we compile a comprehensive finance benchmark consisting of approximately 5.5k multiple-choice questions derived from online exams, LLM-generated questions, and handcrafted instances. This benchmark covers five key topics: financial markets, finance and accounting, domestic company analysis, financial agents~\citep{hu2024infiagent}, and stock price prediction~\citep{soun2022accurate}. Recognizing that multiple-choice questions may not fully represent real-world prompts~\citep{kim2024biggen}, we also include an open-ended QA set featuring 100 challenging prompts.

To encourage the adoption of the benchmark and foster an open research culture, we take a further step by launching an open leaderboard for financial LLMs. It was operated for two months, comprising two stages: a preliminary round and a main round. Over the course of the competition, more than 1,000 models were submitted, with over 600 models remaining publicly accessible to date\footnote{As of 2025.02.28}, thereby laying the groundwork for future research. In addition, we compile the submitted models along with their system cards to document effective tuning strategies. Furthermore, we collect over 200,000 instances from competing teams, filter them, and release a high-quality instruction dataset consisting of 80,000 samples. Finally, after regenerating responses for each instance using Deepseek-R1~\citep{guo2025deepseek} and training on these trajectories, we release \model, the first reasoning model for the Korean financial domain. 

\begin{table*}[!ht]
\centering
\resizebox{\textwidth}{!}{
\begin{tabular}{ll}
\toprule
\textbf{Category} & \textbf{Examples} \\ \midrule

\begin{tabular}[c]{@{}l@{}}Financial Markets \\ \textbf{642 total}\end{tabular} & \begin{tabular}[c]{@{}l@{}}다음 중 대한민국 주식시장 매매 제도에 대한 기술로 알맞은 것은 무엇인가? \\ \textit{\textcolor{gray}{Which of the following descriptions is correct regarding the trading system of the Korean stock market?}} \\ \textit{\textcolor{gray}{A. Opening time is 10:00 AM.}} \\ \textit{\textcolor{gray}{B. The daily price limit for the KOSPI market is \(\pm\)15\% of the previous day's closing price. [...]}} \end{tabular} \\ \midrule

\begin{tabular}[c]{@{}l@{}}Finance and Accounting \\ \textbf{1,450 total}\end{tabular} & \begin{tabular}[c]{@{}l@{}}다음 중 화폐의 시간가치에 관한 설명으로 옳지 않은 것은 무엇인가? \\ \textit{\textcolor{gray}{Which of the following statements about the value of money is incorrect?}} \\ \textit{\textcolor{gray}{A. In monthly compounding, the monthly interest rate is calculated by dividing the annual [...]}} \\ \textit{\textcolor{gray}{B. Given the same initial investment and conditions, compound interest yields higher [...]}} \end{tabular} \\ \midrule

\begin{tabular}[c]{@{}l@{}}Domestic Company Analysis \\ \textbf{2,039 total}\end{tabular} & \begin{tabular}[c]{@{}l@{}}엑세스바이오의 COVID-19 진단 제품의 매출 기여와 미국 시장 판매에 대해서 올바른 것은? \\ \textit{\textcolor{gray}{What is correct regarding the sales contribution of Access Bio’s COVID-19 diagnostic products}} \\ \textit{\textcolor{gray}{and their sales in the U.S. market?}} \\ \textit{\textcolor{gray}{A. Access Bio's COVID diagnostic products were developed for general health screening [...]}} \\ \textit{\textcolor{gray}{B. Access Bio's COVID diagnostic products have demonstrated effectiveness through [...]}} \end{tabular} \\ \midrule

\begin{tabular}[c]{@{}l@{}}Financial Agent \\ \textbf{46 total}\end{tabular} & \begin{tabular}[c]{@{}l@{}}데이터프레임의 ‘종가’ 열의 평균 값을 계산하는 코드를 고르시오. \\ \textit{\textcolor{gray}{Choose the code that calculates the average value of the ‘Closing Price’ column in the DataFrame.}} \\ \textit{\textcolor{gray}{A. df['Close Price'].mean()}} \\ \textit{\textcolor{gray}{B. df['Total Traded Quantity'].median() [...]}} \end{tabular} \\ \midrule

\begin{tabular}[c]{@{}l@{}}Stock Price Prediction \\ \textbf{1,472 total}\end{tabular} & \begin{tabular}[c]{@{}l@{}}주식 A에 대한 분석 결과표를 바탕으로 향후 A의 주가가 상승/하락할지 예측하시오. \\ \textit{\textcolor{gray}{Based on the analysis report of stock A, predict whether the future price of A will rise or fall.}} \end{tabular} \\ \midrule

\begin{tabular}[c]{@{}l@{}} Open-Ended FinQA \\ \textbf{100 total} \end{tabular} & \begin{tabular}[c]{@{}l@{}}위반행위로 얻은 이익이란 무엇이고 그 범위는 어떻게 정의되는가? \\ \textit{\textcolor{gray}{What are the profits gained from breach of contract, and how is their scope defined?}} \end{tabular} \\ \bottomrule

\end{tabular}}
\caption{\footnotesize \textbf{Overview of the benchmark used for evaluation.} Each example demonstrates a specific question type for each category. \textcolor{gray}{Gray} text are English translations provided for better reachability.}
\label{tab:example_error_types}
\end{table*}

\section{Motivation and Related Works}

The financial industry has witnessed rapid expansion in the adoption of artificial intelligence, with particular emphasis on generative AI technologies driving innovations in enhanced customer service, improved risk management, and overall operational efficiency~\citep{mckinsey_2025_ai}.  Despite these advancements, Korean financial institutions face significant challenges in harnessing proprietary language models~\citep{jaech2024openai, team2023gemini}. Strict security regulations—such as network separation policies~\citep{fsc2024misc}—impede their ability to fully leverage these innovations. Moreover, the absence of clear guidelines and robust evaluation frameworks for managing the risks inherent in generative AI—such as hallucinations~\citep{kang2023deficiency}, biases~\citep{zhou2024llms}, and information leakage~\citep{liu2024fmdllama}—further complicates the integration. In response, \citet{son2024krx} introduced \texttt{KRX-Bench}, the first publicly available benchmark designed to assess the knowledge of LLMs in Korean companies. However, \texttt{KRX-Bench} remains limited in scope and has yet to achieve widespread adoption among Korean financial institutions.

In this work, drawing inspiration from financial benchmarks in various languages~\citep{xie2024finben, nie2024cfinbench, koncel2023bizbench}, we extend \texttt{KRX-Bench} to develop a more comprehensive benchmark for Korean financial language models by incorporating five additional categories. Moreover, our work distinguishes itself by operating an open leaderboard with a total prize pool of approximately \$42,000, which has attracted submissions of around 1,000 models, creating the groundwork for future works in financial NLP.


\section{Leaderboard Construction}

In this work, we introduce an open leaderboard for Korean financial LLMs and share lessons learned from its two-month operation, comprised of a preliminary round (October 14–November 7, 2024) and a final round (November 13–December 6, 2024). In total, 1,119 models were submitted—478 in the preliminary round and 641 in the final round, establishing a foundation for open research in Korean financial LLMs, with over 600 models remaining publicly accessible. The following sections describe the benchmark construction process (Section~\ref{s2_details}), present operational details (Section~\ref{s2_operation}) and summarize key statistics (Section~\ref{s2_statistics}).

\subsection{Benchmark Details}\label{s2_details} 

The benchmark used for evaluation consisted of five categories in the preliminary round: finance and accounting, stock price prediction, domestic company analysis, financial markets, and financial agent tasks. For the final round, only three categories were used: finance and accounting, financial markets, and open-ended finance QA. Table~\ref{tab:example_error_types} details the examples of each category.


\paragraph{Finance and Accounting}~For this category, we compile four-option MCQA questions, primarily sourced from university exams. In the preliminary round, these questions were presented with four options, while in the final round, the answer set was expanded to eight options. The augmentation uses two methods: (1) grouping questions based on embeddings to mix similar items, and (2) applying rule-based augmentations~\citep{wang2024mmlu,zhao2024mmlu}, such as replacing an answer option with "none of the above" (thereby making it the correct answer) or shuffling the order of options. A manual human check is done post-augmentation to ensure correctness.

\paragraph{Financial Markets}~For this category, we employ an approach similar to the Finance and Accounting category. However, the source questions are collected from exams that assess understanding of the Korean financial system and related laws.

\paragraph{Stock Price Prediction}~This category is inspired by \citet{soun2022accurate}. We randomly sample fixed-length stock price data (OHLCV: Open, High, Low, Close, Volume) from Korean stock markets, using only post-2024 data to mitigate potential contamination. A set of technical indicators is computed and presented in a Markdown table format (e.g., \texttt{adj-close} for adjusted closing price; \texttt{inc-5}, \texttt{inc-10}, \texttt{inc-15}, \texttt{inc-20}, \texttt{inc-25}, and \texttt{inc-30} for percentage changes over the past 5, 10, 15, 20, 25, and 30 trading days). Models are tasked with a binary classification—predicting whether the price will increase or decrease—and are expected to detect basic signals of momentum~\citep{jegadeesh1993returns} or mean reversion~\citep{poterba1988mean} in the time-series data.

\paragraph{Domestic Company Analysis}~For this section, we directly employ KRX-Bench~\citep{son-etal-2024-krx}, an automatically generated benchmark constructed using GPT-4o~\citep{hurst2024gpt} leveraging annual filings from Korean companies. It consists of 4-option MCQA questions designed to assess knowledge on topics such as Product Offerings, Financial Policy, and Business Strategy.

\paragraph{Financial Agents}~This subset evaluates the capability to function as an automated financial agent by executing code-based tasks on real financial data. Similar to \citet{hu2024infiagent}, the model is provided with a CSV file and an instruction to extract specific information and perform corresponding coding operations. The model is presented with multiple output options, including perturbed variants, and is prompted to select the correct one.

\paragraph{Open-Ended FinQA}~Given that all subsets employ multiple-choice or binary-choice formats, we were concerned that these evaluation methods may not fully capture the diversity of prompts encountered in real-world applications. Drawing inspiration from open-ended evaluations such as MT-Bench~\citep{zheng2023judging}, we curated a set of 100 challenging prompts from three sources: the legal reasoning subset of KRX-Bench~\citep{son2024krx}, advanced math questions from HRM8K~\citep{ko2025understand}, and graduate-level financial engineering and econometrics exam questions. A gold standard answer was generated using o1-Pro~\citep{jaech2024openai}, and GPT-4o was utilized as an LLM-as-a-Judge to determine whether competing models produced responses superior to this standard. Figure~\ref{fig:llm_as_a_judge} illustrates the prompts employed in the LLM-as-a-Judge evaluation.

\subsection{Operation Details} \label{s2_operation}
The leaderboard was active for eight weeks, from October 14, 2024, to November 7, 2024, on a dedicated, self-hosted website. The competition was structured in two rounds: a preliminary round and a final round. In the preliminary round, participants uploaded their models publicly on Hugging Face and submitted the corresponding model links. The top 30 teams advanced to the final round, where each team was allowed up to three submissions. Models were evaluated on a server equipped with 2 A6000 Ada GPUs, with capacity scaling up to 8 GPUs depending on the number of submissions. The benchmark dataset was kept private, with only one sample released from each subset.

To ensure consistency and fairness, participants were restricted in the choice of base models to prevent incompatibility issues with the inference engine and to avoid giving larger companies with more training resources an unfair advantage. Allowed models include Qwen (1.5B and 7B)~\citep{yang2024qwen2}, Mistral (7B)~\citep{jiang2024identifying}, GLM-4 (9B)~\citep{glm2024chatglm}, Llama 3/3.1 (8B)~\citep{grattafiori2024llama}, Amber~\citep{liu2023llm360}, Phi 3.5 (mini)~\citep{abdin2024phi}, and Gemma2 (2B and 9B)~\citep{team2024gemma}. Both base and instruct models were allowed. Teams that advanced to the main rounds were provided \$2500 of AWS credit to help model training.


Participants were required to disclose their datasets and confirm that they did not include any copyrighted material, as a condition for qualifying for the prize money. For evaluation, we adopt a zero-shot chain-of-thought (CoT) format. Initially, each model is prompted to generate a CoT reasoning. We then concatenate the original prompt with the generated CoT and append “\#\#\# Answer:” to prompt the model to produce its final answer. In this step, a logit processor is employed to ensure that the model selects from the provided options, thereby preventing evaluation errors due to format mismatches. To prevent spamming, each team is allowed to submit one model per day.

\subsection{Statistics}\label{s2_statistics}

\paragraph{Submission}~During the preliminary rounds, 233 accounts signed up, with 71 making at least one submission. A total of 478 models were submitted—averaging nearly seven submissions per active team—and November 5 was the busiest day with 45 entries. Moreover, the largest single-day influx of new registrations occurred on October 14, when 83 accounts joined, highlighting strong early interest. For details on the overall trend, see Figure~\ref{fig:sub_stats}. In the main rounds, a high submission rate was maintained throughout the entire period, with 30 teams contributing a total of 641 submissions.

\begin{figure}[h]
    \centering
    \includegraphics[width=\columnwidth]{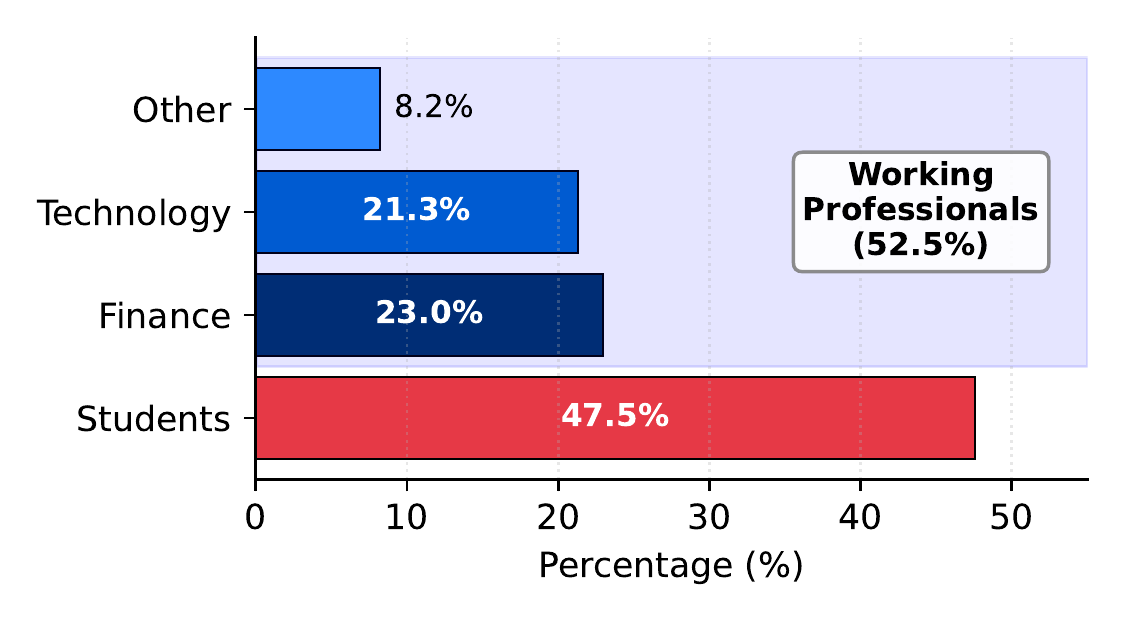}
    \caption{\footnotesize \textbf{Distribution of participants.} The shades of blue bars indicate corporate participants.}
    \label{fig:participants_stats}
    \vspace{-6pt}
\end{figure}

\paragraph{Participants}~A total of 71 teams submitted at least one model during the leaderboard period (each team may consist of up to four members). Among these teams, 52.5\% were corporate participants, with the remaining teams representing universities and student groups. The corporate participants were further categorized into Tech, Finance, and Other sectors, as presented in Figure~\ref{fig:participants_stats}. These results demonstrate that the competition successfully attracted a diverse range of participants—not only students, but also a substantial number of companies, including publicly listed tech companies, securities firms, and banks.

\section{Analysis}

\subsection{Analysis on Data Collection}

As our rules prohibit the use of licensed materials for training LLMs, participants focused on collecting license‐free financial content—potentially useful for constructing corpora to train Korean LLMs. Table~\ref{tab:data_sources} lists the 11 most-used domains, with a strong focus on government (go.kr) and non-profit organizations (or.kr). After collecting the raw corpora from these sources, participants mostly employed either GPT-4o~\citep{hurst2024gpt} or Qwen2.5-72B-Instruct~\citep{yang2024qwen2} to convert the data into MCQA~\citep{bi2024deepseek} or Instruction-Response formats, with some employing an LLM-as-a-Judge~\citep{zheng2023judging, son2024llm} for validation.

\begin{table}[htbp]
\centering
\begin{tabular}{%
  >{\raggedright\arraybackslash}m{2.5cm}  
  >{\raggedright\arraybackslash}m{3.5cm} 
}
\toprule
\textbf{Link} & \textbf{Name} \\
\midrule
\url{krx.co.kr}      & Korea Exchange \\
\url{krxverse.co.kr} & KRXverse \\
\url{fsc.go.kr}      & Financial Services Commission \\
\url{bok.or.kr}      & Bank of Korea \\
\url{law.go.kr}      & \shortstack[l]{Korean Law \\Information Service} \\
\url{kasb.or.kr}     & \shortstack[l]{Korea Accounting \\Standards Board} \\
\url{mss.go.kr}      & \shortstack[l]{Ministry of SMEs\\and Startups} \\
\url{ftc.go.kr}      & Fair Trade Commission \\
\url{kifrs.com}      & K-IFRS \\
\url{kiep.go.kr}     & \shortstack[l]{Korea Institute for\\International Economic\\Policy} \\
\url{kocw.net}       & Korea OpenCourseWare \\
\bottomrule
\end{tabular}
\caption{\textbf{Data collection sources.}}
\label{tab:data_sources}
\end{table}

To ensure reusability, we collect about 200,000 data samples from HuggingFace (released by competing teams) and applied quality filters: the MinHash algorithm to remove near-duplicates, a regex filter to exclude time-bound queries (e.g., “What will Kakao’s 2024 sales be?”), and a rule-based filter to remove incomplete or overly short questions. This process yielded a final set of 86,007 instances. For further details see Appendix~\ref{app_won_instruct}.

\begin{table*}[]
\centering
\fontsize{10}{11}\selectfont
\begin{tabular}{lcccccc}
\multicolumn{1}{c}{\textbf{Models}} & \textbf{F\&A} & \textbf{Stock} & \textbf{Company} & \textbf{Market} & \textbf{Agent} & \textbf{Average} \\
\midrule
AnonymousLLMer/krx-qwen2.5-v1106 & \textbf{\underline{0.51}} & \textbf{\underline{0.56}} & 0.94 & \textbf{\underline{0.49}} & \textbf{\underline{0.83}} & \textbf{\underline{0.67}} \\
AnonymousLLMer/krx-qwen2.5-v1105 & 0.44 & \textbf{\underline{0.56}} & 0.92 & 0.39 & 0.81 & 0.62 \\
KR-X-AI/krx-qwen2-7b-instruct-v4\_m & 0.4 & 0.55 & 0.92 & 0.41 & 0.77 & 0.61 \\
2point5p/krx-qwen2.5-7b-it-prompt-v2 & 0.5 & 0.55 & 0.95 & 0.46 & 0.57 & 0.61 \\
TwoSubPlace/krx-qwen2-7b-it-baseline-v6 & 0.4 & 0.52 & 0.90 & 0.44 & 0.79 & 0.61 \\
KR-X-AI/krx-qwen2-7b-instruct-v3 & 0.4 & 0.49 & 0.9 & 0.44 & 0.72 & 0.59 \\
SejongKRX/Sejong-Qwen-v1 & 0.41 & 0.45 & 0.93 & 0.42 & 0.66 & 0.57 \\
2point5p/krx-qwen2.5-7b-it-X-Two & 0.44 & 0.5 & \textbf{\underline{0.96}} & 0.41 & 0.53 & 0.57 \\
lsw0570168/krx-q25-7b-it-v8 & 0.41 & 0.55 & 0.85 & 0.43 & 0.62 & 0.57 \\
SejongKRX/Sejong-Qwen-v7 & 0.35 & 0.45 & 0.95 & 0.44 & 0.6 & 0.56 \\
\bottomrule
\end{tabular}
\caption{\footnotesize \textbf{Performance of Top-10 models from the preliminary rounds.} The highest performance of each subset is highlighted in \textbf{bold} and the second best is \underline{underlined}.}
\label{tab:top_10}
\end{table*}

\begin{table*}[]
\centering
\fontsize{10}{11}\selectfont
\begin{tabular}{lcccc}
\multicolumn{1}{c}{\textbf{Models}} & \textbf{F\&A} & \textbf{Market} & \textbf{Open-Ended} & \textbf{Average} \\
\midrule
\href{https://huggingface.co/overfit-brothers/hello_world06}{overfit-brothers/hello\_world06} & 0.65 & \textbf{\underline{0.83}} & 0.01 & 0.50 \\
\href{https://huggingface.co/AnonymousLLMer/krx-qwen2.5-v1206-1}{AnonymousLLMer/krx-qwen2.5-v1206-1} & 0.63 & 0.65 & 0.04 & 0.44 \\
\href{https://huggingface.co/shibainu24/qwen2.5-7B-inst-release-1516wk-1519}{shibainu24/qwen2.5-7B-inst-release-1516wk} & 0.56 & 0.67 & 0.04 & 0.43 \\
\href{https://huggingface.co/Q-PING/krx_1205_test_model_3}{Q-PING/krx\_1205\_test\_model\_3} & 0.58 & 0.64 & 0.02 & 0.42 \\
\href{https://huggingface.co/Hi-Q/krx_1206_test_model_2}{Hi-Q/krx\_1206\_test\_model\_2} & 0.60 & 0.61 & 0.02 & 0.41 \\ \midrule
\model (Ours) & \textbf{\underline{0.78}} & 0.66 & \textbf{\underline{0.18}} & \textbf{\underline{0.54}} \\ 
\bottomrule
\end{tabular}
\caption{\footnotesize \textbf{Performance of top 5 models from the main rounds and \model.} \model shows the best average performance with notable improvements in Financial \& Accounting and Open-Ended FinQA.  The highest performance of each subset is highlighted in \textbf{bold} and the second best is \underline{underlined}.}
\label{tab:top_5e}
\end{table*}

\subsection{Analysis on Top-Performing Models}

\begin{figure}[h]
    \centering
    \includegraphics[width=\columnwidth]{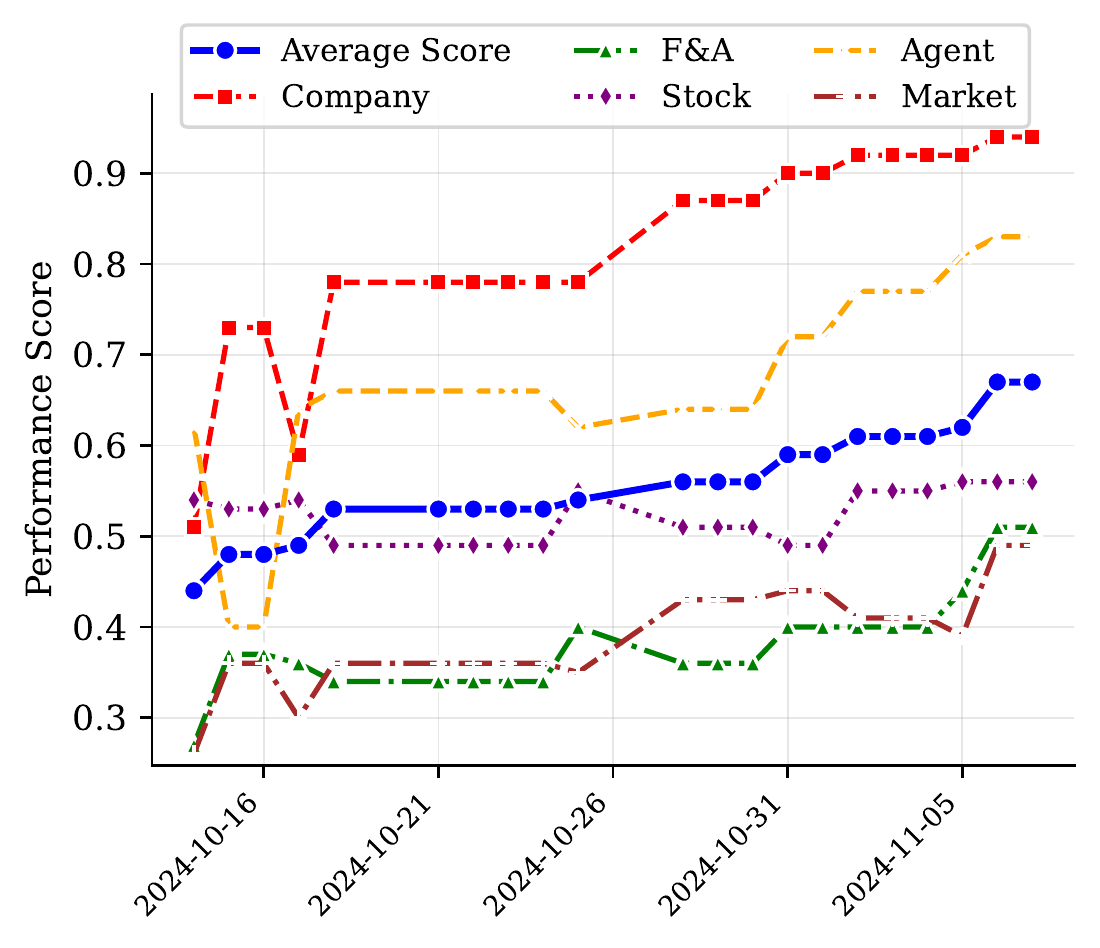}
    \caption{\footnotesize \textbf{Preliminary round performance trends.} }
    \label{fig:perf_stats}
    \vspace{-6pt}
\end{figure}

Table~\ref{tab:top_10} presents the performance of the top 10 models from the preliminary rounds, and Figure~\ref{fig:perf_stats} displays the corresponding score trends. The largest improvement was observed in Domestic Company Analysis, where scores rose from 0.51 to 0.94. However, Financial \& Accounting and Financial Markets experienced relatively modest gains. We attribute this to the relatively simple methods used by most teams during the preliminary rounds. All top 10 teams primarily employed supervised fine-tuning (SFT) for model training. Interstingly, team \texttt{Americano} incorporated a brief continual pre-training phase~\citep{xie2024efficient} before SFT on 3.7GB of text; however, the performance impact of this additional step remains inconclusive in a small-scale setting. Notably, all top-performing models were based on Qwen2.5-7B-Instruct.

Teams advancing to the main rounds employed multi-step, more complex training methods. For example, \texttt{Shinbainu} used a curriculum-based SFT approach that began with training on easier samples and then proceeded to a second round of SFT on more challenging prompts generated via the Evolve Instruct method~\citep{xu2023wizardlm, luo2023wizardmath, luo2023wizardcoder}. The final model was subsequently refined using DPO~\citep{rafailov2023direct}, leveraging preference data from the stage-two SFT model—which generated two responses that were then evaluated by an LLM-as-a-Judge~\citep{zheng2023judging}. Similar strategies were observed among other teams; for instance, \texttt{Hi-Q} and \texttt{Overfit Brothers} implemented KTO~\citep{ethayarajh2024kto} and DPO, respectively.

\begin{figure}
    \centering
    \includegraphics[width=\columnwidth]{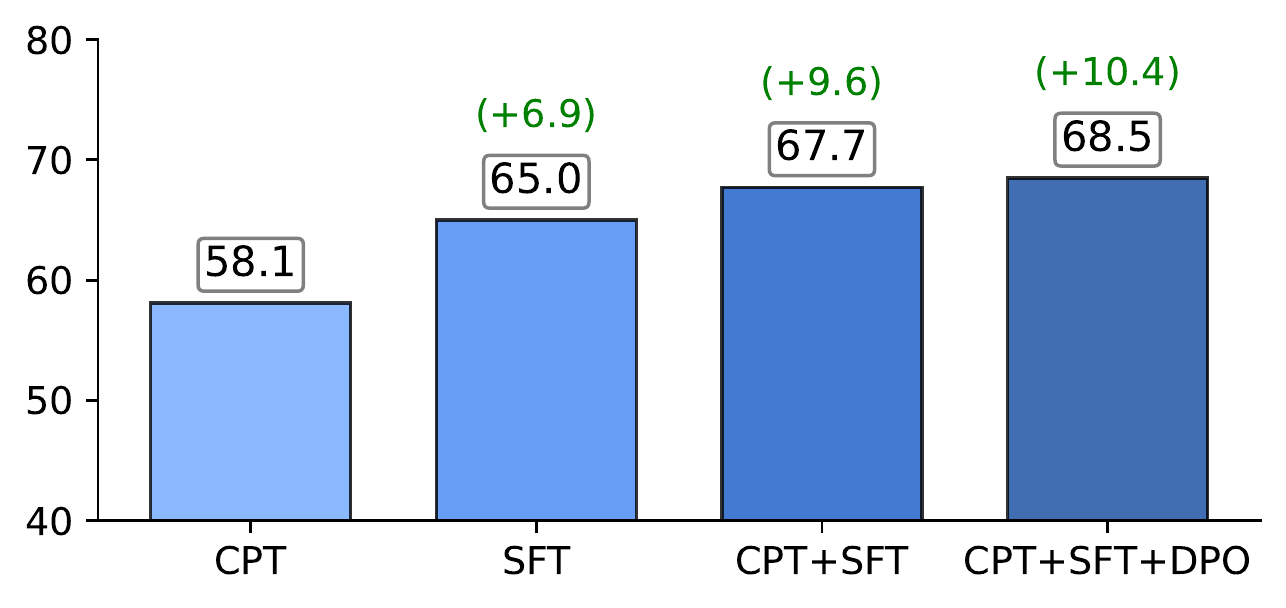}
    \caption{\footnotesize \textbf{Evaluation results reported but \texttt{Hi-Q}.} Performance of each methodology is represented by boxed numbers, and green numbers indicate the improvement over CPT.}
    \label{fig:cpt_result}
\end{figure}

Interestingly, team \texttt{Hi-Q} adopts continual pre-training and demonstrated its effectiveness on a private finance benchmark, as shown in Figure~\ref{fig:cpt_result}. Notably, CPT+SFT scores an average of 2.7 points higher than plain SFT, indicating that a well-structured continued pre-training approach can benefit LLMs in Korean finance. However, further research is required to establish \emph{what} marks a good continual pre-training. Details of the benchmarks used by \texttt{Hi-Q} are provided in Appendix~\ref{app:add_bench}.

\section{\model: Open LLM for Korean Finance}

To aggregate the open resources collected during the competition, we train our own LLM, \model. In line with recent trends toward reasoning LLMs~\citep{jaech2024openai, guo2025deepseek}, \model is designed to generate a two-step response: a first step enclosed within \texttt{<think>} and \texttt{</think>} tags, where the model performs self-correcting reasoning, and a second step enclosed within \texttt{<solution>} and \texttt{</solution>} tags, which provides the final summary of the reasoning process. It should be noted that this effort is not intended to achieve state-of-the-art language model; rather, it serves to evaluate the quality of the collected resources and provide guidelines for future research.

\subsection{Details in Training \model}\label{won_training}

Recent studies have shown that supervised fine-tuning (SFT) is effective enough in training reasoning language models~\citep{muennighoff2025s1, ye2025limo, wen2025light, sun2025tinyr1}. Moreover, during the competition, submissions that have combined SFT with preference optimization techniques such as DPO or KTO have successfully adapted models for the Korean financial domain. Accordingly, we adopt a two-stage training approach: SFT followed by DPO. The SFT dataset comprises prompts paired with responses generated by Deepseek-R1, split evenly between English and Korean. For Korean prompts, the solutions are translated into Korean while retaining the reasoning process in English. The dataset is drawn from three sources: (1) English Prompt-R1 responses collected online~\citep{AM-DeepSeek-R1-Distilled-1.4M}, (2) Korean Prompt-R1 responses collected online~\citep{son2025linguistic}, and (3) 86k prompts from Section 3.1, for which we generated responses using R1. We employed GPT-4o to filter correct samples, retrying up to six attempts for incorrect samples, resulting in approximately 400k instances. Post-SFT, the model struggled with everyday prompts, tended to overthink~\citep{kumar2025overthinking}, and occasionally displayed formatting issues by treating some queries as if they were MCQA tasks. We attribute these issues to the data distribution, which heavily emphasized academic multiple-choice questions paired with extended reasoning. To address these behaviors, we conducted a final DPO stag, where chosen samples are generated from R1, and rejected samples are drawn from the SFT model.

\subsection{Performance Analysis}

The performance of \model is reported in Table~\ref{tab:top_5e}. \model demonstrates strong results in the Finance \& Accounting category, which includes a diverse range of accounting and econometrics tasks that benefit from robust mathematical and logical reasoning. These capabilities also yield strong performance on open-ended FinQA tasks, where multi-step logical deductions are necessary. In contrast, \model shows weaker performance in the Market category. The Market category relies more on factual and domain-specific knowledge; they do not benefit as strongly from \model's reasoning-oriented approach. These findings are consistent with those of \citet{ha2025pensezdatabetterreasoning}, who observed that while reasoning-focused models excel at challenging mathematical questions, their performance decline in knowledge-intensive domains as training progresses.

\section{Conclusion}

In this work, we present the largest Korean finance benchmark covering five categories: finance and accounting, stock price prediction, domestic company analysis, financial markets, and financial agent tasks. To encourage adoption, we launched a leaderboard that attracted hundreds of participants from academia and industry, resulting in around 600 publicly available models. We distilled successful strategies from these submissions into an 80k-instruction dataset, which we used to train and release \model, a publicly available reasoning model for Korean finance.

\bibliography{custom}

\newpage
\appendix
\onecolumn

\section{Further details on \dataset}\label{app_won_instruct}

Here we report the average length of questions and responses using the Unimax tokenizer proposed by ~\citet{chung2023unimax}. 

\begin{figure*}[htbp]
    \centering
    \begin{subfigure}[b]{0.45\linewidth}
        \centering
        \includegraphics[width=\linewidth]{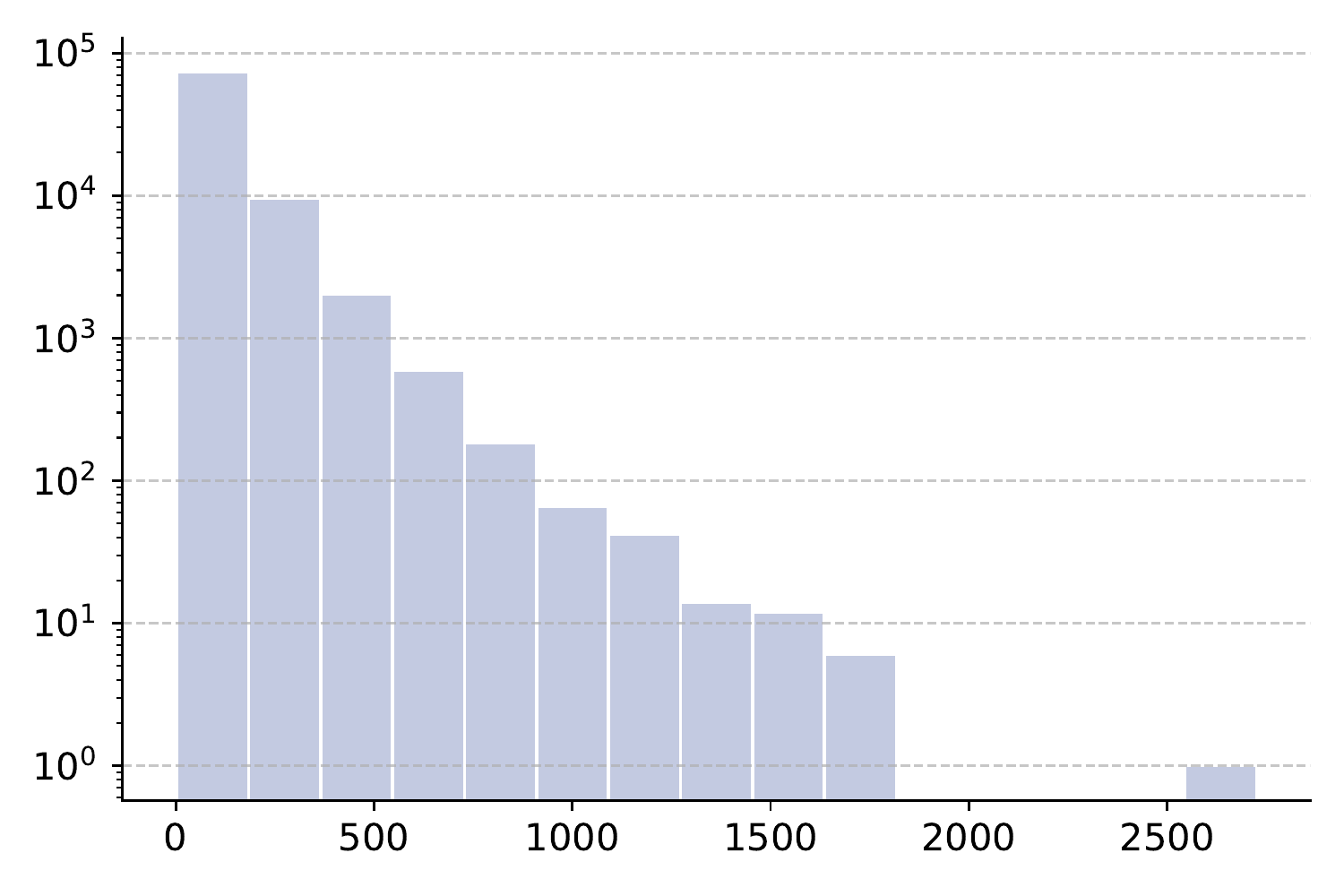}
        \caption{Prompt length}
    \end{subfigure}
    \hfill
    \begin{subfigure}[b]{0.45\linewidth}
        \centering
        \includegraphics[width=\linewidth]{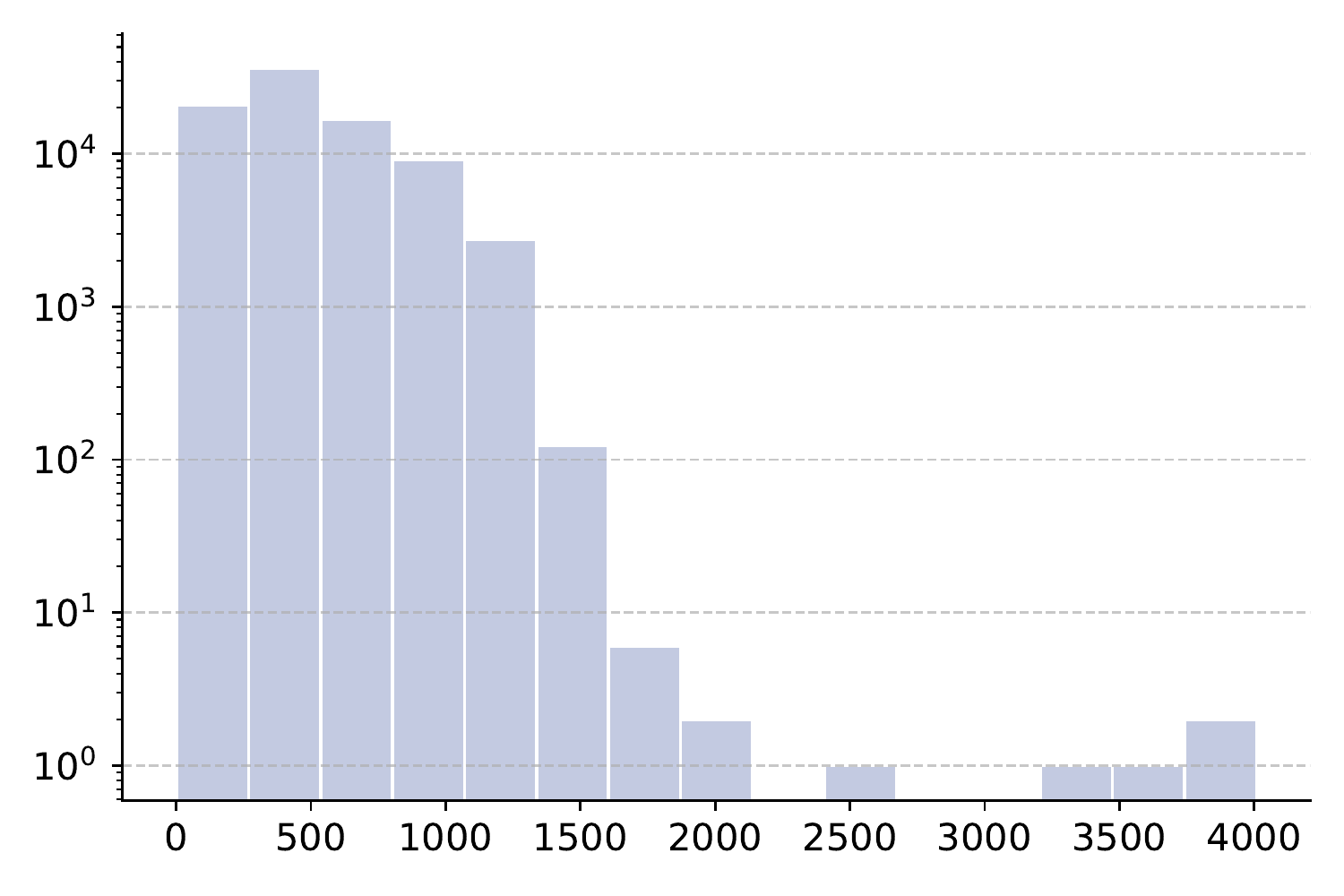}
        \caption{Response length}
    \end{subfigure}

    \caption{\textbf{Statistics of prompt and response length in \dataset.}}
    \label{fig:dataset_statistics}
\end{figure*}

\section{Training details for \model}

Axolotl~\citep{axolotl2025} is used for the SFT and DPO training in Section~\ref{won_training}. We train Qwen2.5-Math-7B-Instruct with DeepSpeed-Zero1~\citep{rajbhandari2020zero} on 8 H100 80GB GPUs for 25 hours. \citet{hsu2024liger} is used for optimization. Table~\ref{tab:training_configs_1} and \ref{tab:training_configs_2} are configurations used for SFT and DPO respectively.

\begin{table}[hb]
\centering
\begin{minipage}{0.45\linewidth}
\centering
\fontsize{9}{11}\selectfont
\begin{tabular}{cc}
\toprule
\textbf{Category} & \textbf{Section~\ref{won_training}} \\ \midrule
\textbf{Sequence Length} & 16,384 \\
\textbf{Learning Rate} & \(4 \times 10^{-5}\) \\
\textbf{Global Batch (Effective)} & 256 \\
\textbf{Learning Rate Scheduler} & Cosine Decay \\
\textbf{Warmup Ratio} & 0.05 \\
\textbf{Training Epochs} & 2 \\ \bottomrule
\end{tabular}
\caption{\footnotesize \textbf{SFT configuration details for Section~\ref{won_training}.}}
\label{tab:training_configs_1}
\end{minipage}\hfill
\begin{minipage}{0.45\linewidth}
\centering
\fontsize{9}{11}\selectfont
\begin{tabular}{cc}
\toprule
\textbf{Category} & \textbf{Section~\ref{won_training}} \\ \midrule
\textbf{Sequence Length} & 16,384 \\
\textbf{Learning Rate} & \(5 \times 10^{-6}\) \\
\textbf{Global Batch (Effective)} & 64 \\
\textbf{Learning Rate Scheduler} & Cosine Decay \\
\textbf{Warmup Ratio} & 0.05 \\
\textbf{Training Epochs} & 1 \\ \bottomrule
\end{tabular}
\caption{\footnotesize \textbf{DPO configuration details for Section~\ref{won_training}.}}
\label{tab:training_configs_2}
\end{minipage}
\end{table}

\section{Further details on private evaluation tools used by team \texttt{Hi-Q} }\label{app:add_bench}

In Figure~\ref{fig:cpt_result}, we share private evaluation results conducted by \texttt{Hi-Q}. The evaluation is done on a private benchmark consisting of financial questions collected from sources such as AiHub\footnote{\url{https://www.aihub.or.kr/}} and KMMLU~\citep{son2024kmmlu}, to assess the model's financial knowledge and capability. In particular, the private benchmark comprises the following subsets:
\begin{itemize}
    \item \textbf{Accounting}: A private question set on Korean accounting.
    \item \textbf{Financial Accounting Generated}: Synthetically generated using GPT-4 on sample instances, following a \citet{wang2022self}-like approach (also applied to the Financial Market Generation subset).
    \item \textbf{KMMLU-accounting}: The accounting subset of the KMMLU dataset.
    \item \textbf{AiHUB-NC-MRC}: A dataset provided by AiHUB focusing on numerical computation and machine reading comprehension~\citep{aihub_dataset_71568}.
    \item \textbf{AiHUB-FL-MRC)}: A dataset provided by AiHUB focusing on financial and law machine reading comprehension~\citep{aihub_dataset_71610}.
\end{itemize}

The benchmark evaluation results of methodologies attempted by \texttt{Hi-Q} are presented in Table~\ref{tab:additional_benchmark}.

\begin{table*}[ht!]
\centering
\fontsize{9}{11}\selectfont
\begin{tabular}{clcccc}
\toprule
\textbf{Category} & \multicolumn{1}{c}{\textbf{Subset}} & \textbf{CPT} & \textbf{SFT} & \textbf{CPT+SFT} & \textbf{CPT+SFT+DPO} \\ \midrule
\multirow{6}{*}{\makecell{Financial \\ Accounting}} & Accounting & 32.0 & 39.0 & 41.0 & \textbf{43.0} \\
 & Financial\_Accounting\_Generated & 55.0 & 71.0 & 70.0 & \textbf{73.0} \\
 & KMMLU\_Accounting & 37.0 & 42.0 & 41.0 & \textbf{44.0} \\
 & AiHUB-NC-MRC\_calculation & 55.0 & 57.0 & 60.0 & \textbf{61.0} \\
 & AiHUB-NC-MRC\_boundary\_extraction & 85.0 & 91.0 & \textbf{95.0} & \textbf{95.0} \\
 & AiHUB-NC-MRC\_multilateral\_comparison & 50.0 & 49.0 & \textbf{59.0} & 56.0 \\ \midrule
\multirow{4}{*}{Financial Markets} & AiHUB-FL-MRC\_mcqa & 52.0 & \textbf{67.0} & 66.0 & 62.0 \\
 & AiHUB-FL-MRC\_process & 80.0 & 84.0 & 84.0 & \textbf{89.0} \\
 & AiHUB-FL-MRC\_answer\_boundary & 83.0 & 90.0 & \textbf{94.0} & 93.0 \\
 & Financial\_Market\_Generated & 52.0 & 60.0 & 67.0 & \textbf{69.0} \\ \midrule
\multicolumn{2}{c}{\textbf{Avg.}} & 58.1 & 65.0 & 67.7 & \textbf{68.5} \\ \bottomrule
\end{tabular}
\caption{The internal benchmark results of \texttt{Hi-Q}. The \textbf{bold} font indicates that the highest score of each section.}
\label{tab:additional_benchmark}
\end{table*}

\section{Additional resources}

\begin{figure}[h]
    \centering
    \includegraphics[width=0.5\columnwidth]{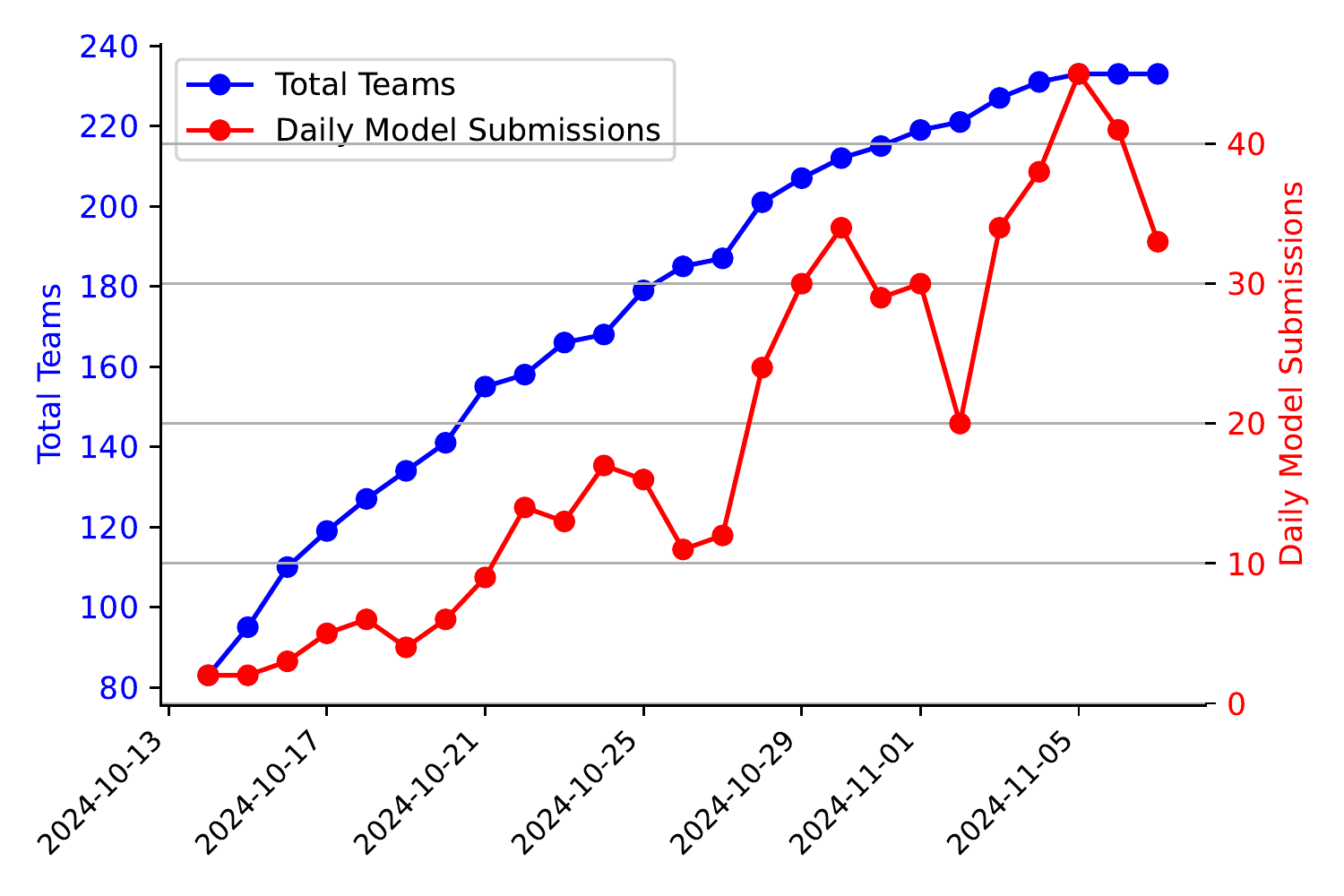}
    \caption{Model submission trends during the preliminary rounds.}
    \label{fig:sub_stats}
\end{figure}

In this section, we present additional resources that were excluded from the main text due to space constraints:
\begin{enumerate}
    \item Figure~\ref{fig:sub_stats}: Model submission trends during preliminary rounds from Section~\ref{s2_statistics}.
    \item Figure~\ref{fig:llm_as_a_judge}: Sample prompt used for LLM-as-a-Judge to evaluate the Open-Ended FinQA subset from Section~\ref{s2_details} .
\end{enumerate}

\begin{figure*}[h]
\begin{instructionsbox}
\begin{lstlisting}
[System]
Please act as an impartial judge and evaluate the quality of the responses provided by two AI assistants to the user question displayed below. You should choose the assistant that follows the user's instructions and answers the user's question better. Your evaluation should consider factors such as the helpfulness, relevance, accuracy, depth, creativity, and level of detail of their responses. Begin your evaluation by comparing the two responses and provide a short explanation. Avoid any position biases and ensure that the order in which the responses were presented does not influence your decision. Do not allow the length of the responses to influence your evaluation. Do not favor certain names of the assistants. Be as objective as possible. After providing your explanation, output your final verdict by strictly following this 
format: "[[A]]"  if assistant A is better, "[[B]]" if assistant B is better, and "[[C]]" for a tie.

[User Question]
{question}

[The Start of Assistant A's Answer]
{answer_a}
[The End of Assistant A's Answer]

[The Start of Assistant B's Answer]
{answer_b}
[The End of Assistant B's Answer]
\end{lstlisting}
\end{instructionsbox}
\caption{Prompt used for LLM-as-a-Judge to evaluate the Open-Ended FinQA subset}
\label{fig:llm_as_a_judge}
\end{figure*}

\section{Additional Results}\label{app_results}

In Tables~\ref{tab:preliminary_perf} and \ref{main_perf}, we present the performance of baseline models on the benchmarks used for the preliminary and main rounds, respectively. The tables include results for Qwen (1.5B and 7B)\citep{yang2024qwen2}, Mistral (7B)\citep{jiang2024identifying}, GLM-4 (9B)\citep{glm2024chatglm}, Llama 3/3.1 (8B)\citep{grattafiori2024llama}, Amber~\citep{liu2023llm360}, Phi 3.5 (mini)\citep{abdin2024phi}, and Gemma2 (2B and 9B)\citep{team2024gemma}.

\begin{table*}[ht]
\centering
\fontsize{6}{8}\selectfont
\resizebox{\textwidth}{!}{%
\begin{tabular}{lccccc|c}
\textbf{Models} & \textbf{Company} & \textbf{F\&A} & \textbf{Stock} & \textbf{Agent} & \textbf{Market} & \textbf{Average} \\
\midrule
Qwen2-7B-Instruct & 0.51 & 0.27 & 0.54 & 0.62 & 0.26 & 0.44 \\
gemma-2-9b & 0.31 & 0.25 & 0.54 & 0.43 & 0.27 & 0.36 \\
Llama-3.1-8B & 0.40 & 0.22 & 0.56 & 0.38 & 0.22 & 0.36 \\
gemma-2-9b-it & 0.39 & 0.28 & 0.55 & 0.32 & 0.25 & 0.36 \\
Qwen2-7B & 0.29 & 0.21 & 0.55 & 0.45 & 0.25 & 0.35 \\
Llama-3.2-3B & 0.43 & 0.23 & 0.55 & 0.32 & 0.20 & 0.35 \\
Qwen2.5-7B & 0.33 & 0.24 & 0.54 & 0.40 & 0.24 & 0.35 \\
Meta-Llama-3-8B & 0.38 & 0.23 & 0.56 & 0.30 & 0.23 & 0.34 \\
Qwen2.5-1.5B-Instruct & 0.27 & 0.26 & 0.54 & 0.30 & 0.21 & 0.34 \\
Qwen2.5-3B & 0.37 & 0.22 & 0.54 & 0.28 & 0.22 & 0.33 \\
Qwen2.5-7B-Instruct & 0.32 & 0.28 & 0.51 & 0.34 & 0.22 & 0.33 \\
Mistral-7B-Instruct-v0.3 & 0.37 & 0.24 & 0.54 & 0.30 & 0.21 & 0.33 \\
Llama-3.2-3B-Instruct & 0.27 & 0.23 & 0.50 & 0.40 & 0.20 & 0.33 \\
Qwen2.5-3B-Instruct & 0.30 & 0.25 & 0.54 & 0.28 & 0.22 & 0.32 \\
Llama-3.1-8B-Instruct & 0.28 & 0.25 & 0.51 & 0.32 & 0.24 & 0.32 \\
Qwen2.5-1.5B & 0.30 & 0.25 & 0.56 & 0.26 & 0.23 & 0.31 \\
Qwen2-1.5B-Instruct & 0.26 & 0.22 & 0.53 & 0.33 & 0.24 & 0.31 \\
gemma-2-2b & 0.25 & 0.23 & 0.55 & 0.32 & 0.19 & 0.31 \\
Llama-3.2-1B & 0.27 & 0.26 & 0.55 & 0.23 & 0.18 & 0.30 \\
Mistral-7B-Instruct-v0.2 & 0.28 & 0.21 & 0.54 & 0.26 & 0.23 & 0.30 \\
gemma-2-2b-it & 0.32 & 0.24 & 0.49 & 0.28 & 0.18 & 0.30 \\
Qwen2-1.5B & 0.25 & 0.20 & 0.54 & 0.30 & 0.22 & 0.30 \\
Llama-3.2-1B-Instruct & 0.28 & 0.23 & 0.52 & 0.21 & 0.19 & 0.29 \\
AmberChat & 0.26 & 0.23 & 0.53 & 0.23 & 0.21 & 0.29 \\
Amber & 0.25 & 0.23 & 0.54 & 0.23 & 0.21 & 0.29 \\
Meta-Llama-3-8B-Instruct & 0.28 & 0.24 & 0.53 & 0.21 & 0.21 & 0.29 \\
Mistral-7B-Instruct-v0.1 & 0.22 & 0.21 & 0.55 & 0.28 & 0.22 & 0.29 \\
Mistral-7B-v0.3 & 0.27 & 0.20 & 0.51 & 0.23 & 0.21 & 0.28 \\
Phi-3.5-mini-instruct & 0.25 & 0.25 & 0.54 & 0.17 & 0.18 & 0.28 \\
Mistral-7B-v0.1 & 0.29 & 0.20 & 0.53 & 0.17 & 0.21 & 0.28 \\
\bottomrule
\end{tabular}%
}
\caption{\textbf{Performance of base models in preliminary round}}
\label{tab:preliminary_perf}
\end{table*}

\begin{table*}[h]
\centering
\fontsize{11}{12.5}\selectfont
\begin{tabular}{lccc|c}
\textbf{Models} & \textbf{F\&A} & \textbf{Market} & \textbf{Open-Ended} & \textbf{Average} \\
\midrule
gemma-2-9b-it & 0.43 & 0.64 & 0.00 & 0.36 \\
Qwen2.5-7B-Instruct & 0.50 & 0.56 & 0.00 & 0.35 \\
Qwen2-7B-Instruct & 0.45 & 0.53 & 0.00 & 0.33 \\
Qwen2.5-3B-Instruct & 0.40 & 0.52 & 0.00 & 0.31 \\
Qwen2.5-7B & 0.37 & 0.41 & 0.00 & 0.28 \\
Meta-Llama-3-8B-Instruct & 0.37 & 0.43 & 0.00 & 0.27 \\
Qwen2-7B & 0.37 & 0.40 & 0.00 & 0.26 \\
Llama-3.1-8B-Instruct & 0.32 & 0.45 & 0.00 & 0.26 \\
Phi-3.5-mini-instruct & 0.38 & 0.36 & 0.00 & 0.25 \\
gemma-2-9b & 0.32 & 0.41 & 0.00 & 0.24 \\
Qwen2.5-1.5B-Instruct & 0.34 & 0.35 & 0.00 & 0.23 \\
Qwen2.5-3B & 0.30 & 0.36 & 0.00 & 0.22 \\
Mistral-7B-Instruct-v0.3 & 0.32 & 0.30 & 0.00 & 0.21 \\
Llama-3.1-8B & 0.24 & 0.36 & 0.00 & 0.20 \\
Llama-3.2-3B-Instruct & 0.26 & 0.34 & 0.00 & 0.20 \\
Qwen2-1.5B-Instruct & 0.20 & 0.28 & 0.00 & 0.19 \\
Qwen2.5-1.5B & 0.27 & 0.28 & 0.00 & 0.18 \\
Meta-Llama-3-8B & 0.25 & 0.30 & 0.00 & 0.18 \\
gemma-2-2b-it & 0.22 & 0.32 & 0.00 & 0.18 \\
gemma-2-2b & 0.30 & 0.23 & 0.00 & 0.18 \\
Mistral-7B-Instruct-v0.1 & 0.30 & 0.22 & 0.00 & 0.17 \\
Mistral-7B-v0.1 & 0.24 & 0.26 & 0.00 & 0.17 \\
Llama-3.2-3B & 0.24 & 0.25 & 0.00 & 0.16 \\
Llama-3.2-1B & 0.24 & 0.25 & 0.00 & 0.16 \\
AmberChat & 0.24 & 0.24 & 0.00 & 0.16 \\
Qwen2-1.5B & 0.22 & 0.25 & 0.00 & 0.16 \\
Mistral-7B-Instruct-v0.2 & 0.22 & 0.24 & 0.00 & 0.15 \\
Mistral-7B-v0.3 & 0.21 & 0.22 & 0.00 & 0.14 \\
Amber & 0.24 & 0.19 & 0.00 & 0.14 \\
Llama-3.2-1B-Instruct & 0.22 & 0.20 & 0.00 & 0.14 \\
\bottomrule
\end{tabular}
\caption{\textbf{Performance of base models in main round}}
\label{main_perf}
\end{table*}

\end{document}